\documentclass[a4paper,10pt]{article}
\usepackage{longtable}

\title{Identification of parameters underlying emotions and a classification of emotions}
\author{N. Arvind Kumar \\ {\small arvind.kumar@gmail.com}}

\begin{document}

\maketitle

\begin{abstract}
 The standard classification of emotions involves categorizing the expression of emotions. In this paper, parameters underlying some emotions are identified and a new classification based on these parameters is suggested.
\end{abstract}

\section{Introduction }

Widely accepted classifications of emotions classify them by the expression of emotions. Paul Ekman has proposed a list of basic emotions and has considered the expression of the emotion as a characteristic of the emotion~\cite{ekman}. The list includes anger, fear, sadness, enjoyment, disgust and surprise. Another classification by Robert Plutchik also classifies emotions based on the external expression of emotions. Marvin Minsky states that ``emotion is one of those suitcase like words that we use to conceal the complexity of very large ranges of different things.''~\cite{minsky}

The approach in this paper does not attempt to identify basic emotions, but is in alignment with Minsky's argument about the complex nature of emotions. It deviates from the method of classifying emotions based on the expression of emotions. Instead, the underlying parameters that result in an emotion are taken into consideration for the purpose of organizing various emotions. One of the biggest challenges in categorizing emotions in this manner is that the brain must observe itself and separate out the various strands that are part of the complex mental process that is observed. It is not an easy task to identify the individual strands of the complex process. Even when the individual strands are identified, it is not easy to determine the particular mental process to which they belong.

\section{Classification of emotions}

Emotions depend on attachments or repulsion (negative attachments) to certain states or desired states and emotions would not exist in the absence of such attachments. There are other parameters apart from an attachment that go into the formation of an emotion. These parameters are used here to differentiate emotions from each other.

In the categorization of emotions made here, emotions are treated as pairs - the elements of an emotion pair have identical parameters underlying them except that one is negative (related to pain) while the other is positive (related to comfort). For example, in the confidence and fear pair, the assessment of the expected future state is common to both the emotions, but one is a positive emotion while the other is a negative emotion. Similarly, surprise and shock result from the observed state being unexpected. The unexpectedness of a state is common to both these emotions and hence they belong to a pair. Another example is the pair containing pride and shame. Both of these depend on knowing whether others are aware of the actions or circumstance of the person in question and whether or not such knowledge by others is desired. If one element of the pair is directed towards something (anger), so too is the other element in the pair. Thus the classification in this paper is such that the elements of each pair maintain some sort of symmetry.

It is also possible to consider the absence of an emotion as the antonym of a particular emotion and pair them accordingly. For example, anger and tranquility could form a pair, but that is not what is intended in this paper. The intended pairing involves a positive emotion and a negative emotion. Absence of an emotion can be achieved by lowering the intensity of the attachment to zero.

Negative emotions arise from negative attachments and positive emotions arise from positive attachments. The intensity of an emotion depends on the intensity of the underlying attachment. Many emotions pairs listed below merely differ in intensity from similar pairs.

Marvin Minsky has proposed a six-level model of mental activities consisting of instinctive reactions, learned reactions, deliberative thinking, reflective thinking, self-reflective thinking and self-conscious emotions~\cite{minsky1}. The emotions listed below too show various levels of complexity and many of the pairs listed here can be mapped onto one or more categories of these six levels although there is no accurate mapping from the emotion pairs to the six-level model. No attempt has been made to group the emotions pairs by these levels although there is a vague relationship between the pairs and the six-level model.

\section{Characteristics of the emotion pairs}

The emotion pairs in the table below begin with names for attachments and become more complex as we go down the table. The first two pairs, repulsion-attraction and hate-love, are names for attachments of different intensities.

The next four pairs are responses to observations but each pair has its own characteristic. While the elements of the sadness-happiness pair evoke a sense of loss and sense of gain, the intended meanings for the discomfort-comfortable and rejection-acceptance pairs is that the reasons for the emotions in the former pair arise subconsciously while the reasons for the emotions in the latter pair are directly observed. Disgust and pleasantness are strong reactions to rejection and acceptance respectively.

The next three pairs consist of emotions that depend on unexpected situations. They not only differ in intensities, but the elements in the horror-wonder pair are mixed with a sense of disgust and admiration respectively. In addition, the nature of unexpectedness in this pair is such that the observed situation was never considered as a possibility, while in the other two pairs, the observed situation might have been considered and rejected as an impossible situation.

The next three pairs depend on the future state. While the emotions in the fear-confidence pair result from an assessment of the future state, the emotions in the dread-eagerness pair are reactions to fear and confidence. Anxiety and anticipation are very mild forms of dread and eagerness, and the elements in the despair-hope pair come into operation when an adverse situation is the most likely outcome.

In the next three pairs, the common strand is an object at which the emotion is directed. While the emotions in the blame-appreciation pair fix responsibility (possibly on an unknown or as yet undefined entity) for the observed situation, the emotions in the next pair, anger and its counterpart, are reactions or at least intentions to react after fixing responsibility for the observed situation. In vengeful-grateful pair, the reaction is stretched out over a period of time and involves analytical thinking.

The next four pairs involve self-reflection. The first of these is humble-haughty which involve assessing one's self-worth (possibly in comparison with others). The elements of the regret-gladness pair are emotions that occur when an unfavorable or favorable event happens. Typically, these include events where one's own actions are involved. In guilt and good conscience, there is also a value system that is involved. The next pair, remorse and its counterpart, consists of a mixture of the elements in the previous two pairs.

In the next three pairs, a third party is involved but the emotion is directed at oneself. Shyness and boastfulness are states where one wants to either avoid others observing oneself or one's characteristics. Embarrassment and a feeling of being honored are created when attention from others is focused on one's negative or positive attributes. Shame and pride are similar to embarrassment and feeling honored except that these involve actions related to one's ideals and values.

Finally, the last six pairs also involve third parties, but the emotions in these pairs are directed at the third parties. Contempt and respect as well as illwill and goodwill are directed at others. The three pairs, pity-sadism, cruelty-kindness and envy-magnanimity can be considered subsets of the illwill-goodwill pair under various conditions of the third party. The greed-generosity pair exists only when a coveted resource is scarce.

\section{Table of emotions}

A challenge in describing emotions is that a word in a particular language may have multiple meanings and these words do not conform to any logical rules. This is true of English too. Careful effort has been made to choose words that describe emotions. The specific meaning of a particular emotion is listed in the table. For the purpose of this classification, all other meanings are excluded even if they are legitimate meanings for the word in common usage. The meaning that is listed here has been chosen from \textit{Concise Oxford Dictionary of Current English, Seventh Edition} except where specified. The first word in a pair describes a negative emotion while the second describes a positive emotion. The word in parenthesis is the actual form of the word that was looked up in the dictionary.

\begin{center}
  \begin{longtable}{l p{6.9cm} l}
    \hline \hline
    {\bf Emotion} &	{\bf Intended meaning of emotion}\\
    \hline \hline
repulsion & aversion \\
attraction & act or faculty of drawing to oneself or itself \\
    \hline
hate & have strong dislike of or strong aversion to; dislike greatly\\
love &  attachment \\
\hline
sadness	& affected with or expressive of grief or unhappiness (from m-w.com) \\
happiness & a pleasurable and satisfying experience (from m-w.com) \\
\hline
discomfort & uneasiness of body or mind \\
 comfortable & at ease \\
\hline
rejection (reject) & refuse acceptance of \\
acceptance & approval, toleration\\
\hline
disgust & repugnance, strong aversion \\
pleasantness (pleasant) & agreeable to mind, feelings, or senses\\
\hline
shock & sudden and disturbing physical or mental impression \\
surprise & emotion excited by the unexpected\\
\hline
horror & shock and painful feeling of loathing and fear \\
wonder & emotion excited by what surpasses expectation or experience or what seems inexplicable, surprise mingled with admiration or curiosity or bewilderment\\
\hline
stun & to overcome especially with paralyzing astonishment or disbelief \\
amazement (amaze) & overwhelm with wonder \\
\hline
fear & apprehend, have uneasy expectation of \\
confidence & assured expectation\\
\hline
dread & look forward to with terror, be afraid; fear greatly (that, to learn etc.) \\
eagerness & full of keen desire\\
\hline
anxiety & concern about the future \\
anticipation (anticipate) & look forward to\\
\hline
despair & complete loss or absence of hope \\
hope & expectation and desire combined\\
\hline
blame & find fault with; fix the responsibility on \\
appreciation & adequate recognition\\
\hline
angry & extremely displeased, resentful (at, about, thing; at, with, person;) \\
pleased with & derive pleasure or satisfaction from\\
\hline
vengeful & disposed to revenge \\
grateful & feeling or showing gratitude (to person, for thing; that, to do)\\
\hline
humble & having or showing low estimate of one's own importance \\
haughty & valuing oneself too highly\\
\hline
regret & be distressed about or sorry for (event, fact)\\
gladness & (of news or event) giving joy\\
\hline
guilt & mental obsession with idea of having done wrong \\
good conscience & see definitions of gladness and good conscience\\
\hline
shyness (shy) & avoiding observation; diffident or uneasy in company \\
boastful & praise oneself; make boasts of\\
\hline
embarrassed (embarrass) & make (person) feel awkward or ashamed \\
honored (honor) & confer dignity upon\\
\hline
shame & feeling of humiliation excited by consciousness of (esp. one's own) guilt or shortcoming \\
pride & feeling of elation and pleasure due to action or circumstance that does one credit\\
\hline
contempt & act or mental attitude of despising \\
respect & regard with deference, esteem or honor\\
\hline
illwill & hostile, unkind \\
goodwill & kindly feeling to person, virtuous intent\\
\hline
cruelty & having or showing indifference to or pleasure in another's suffering; causing pain or suffering \\
kindness (kind) & of gentle or benevolent nature; friendly in one's conduct to (person etc)\\
\hline
pity & feeling of sorrow aroused by person's distress or suffering \\
sadism & deriving of pleasure from inflicting or watching cruelty\\
\hline
envy & resentful or admiring contemplation (of more fortunate person, of, at, his advantages, or abs.)\\
magnanimity (magnanimous) & noble, generous, not petty, in feelings or conduct\\
\hline
greed (greedy) & avaricious, covetous; intensely desirous \\
generosity & free in giving, munificent\\

\hline
  \end{longtable}
\end{center}

\section{Conclusion}

A new classification of emotions by grouping them into pairs based on certain mental processes underlying these emotions has been proposed. This method ignores the external expression of emotions completely. Elements in each pair are symmetrical with respect to each other in the sense that they contain identical sets of parameters that underlie them except that one element is a negative emotion while the other is a positive emotion. This classification uses these underlying parameters of emotions instead of treating emotions as black boxes. It will be particularly useful for those who want to model emotions in the field of artificial intelligence.

\end{document}